\pdfoutput=1

\documentclass[11pt]{article}

\usepackage[]{ACL2023}

\usepackage{times}
\usepackage{latexsym}

\usepackage[T1]{fontenc}

\usepackage[utf8]{inputenc}

\usepackage{microtype}

\usepackage{inconsolata}

\usepackage{graphicx}
\usepackage{amsmath}
\usepackage{multirow}

\title{Comprehensive Solution Program Centric Pretraining for Table-and-Text Hybrid Numerical Reasoning}

\author{Qianying Liu$^1$, Dongsheng Yang$^1$, Wenjie Zhong$^2$, Fei Cheng$^1$ and Sadao Kurohashi$^1$\\
$^1$ Graduate School of Informatics, Kyoto University \\
$^2$ The University of Tokyo\\
  {\tt ying@nlp.ist.i.kyoto-u.ac.jp; 
  yang.dongsheng.46w@st.kyoto-u.ac.jp; zvengin@is.s.u-tokyo.ac.jp} \\
  \tt \{feicheng,kuro\}@nlp.ist.i.kyoto-u.ac.jp;
  }
  
\begin{document}
\maketitle
\begin{abstract}

Numerical reasoning over table-and-text hybrid passages, such as financial reports, poses significant challenges and has numerous potential applications. Noise and irrelevant variables in the model input have been a hindrance to its performance. Additionally, coarse-grained supervision of the whole solution program has impeded the model's ability to learn the underlying numerical reasoning process. In this paper, we propose three pretraining tasks that operate at both the whole program and sub-program level: Variable Integrity Ranking, which guides the model to focus on useful variables; Variable Operator Prediction, which decomposes the supervision into fine-grained single operator prediction; and Variable Keyphrase Masking, which encourages the model to identify key evidence that sub-programs are derived from. Experimental results demonstrate the effectiveness of our proposed methods, surpassing transformer-based model baselines.

\end{abstract}

\section{Introduction}

\begin{figure*}[t]
  \centering
  \includegraphics[width=\linewidth]{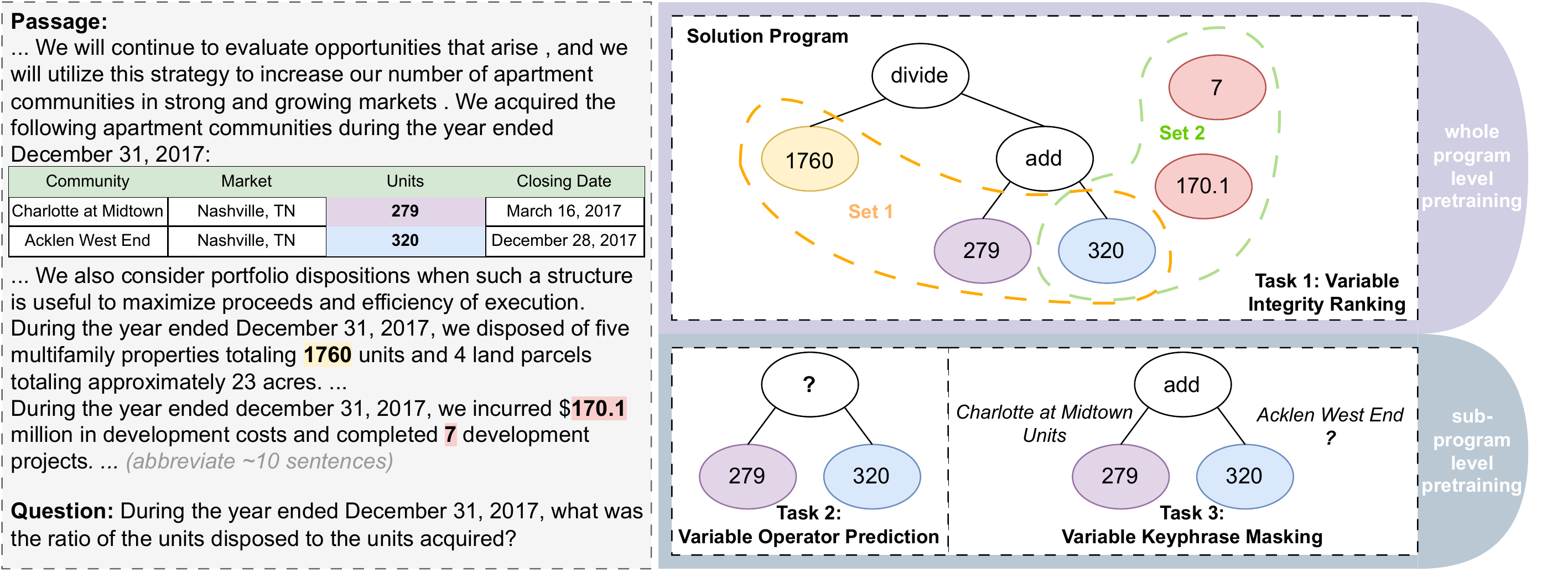}
  \caption{Example of Hybrid Numerical QA and our proposal. The yellow, purple and blue color denotes required variables for solving the question. The red variables denotes irrelavant variables.}
  \label{fig:intro}
  \end{figure*}

The field of natural language processing has seen a growing interest in developing techniques for Question Answering (QA) style numerical reasoning on both textual data~\cite{huang-etal-2016-well, wang-etal-2017-deep, dua-etal-2019-drop} and tabular data~\cite{cheng-etal-2022-hitab}. Recent research has focused on addressing the challenge of numerical reasoning over Table-and-Text hybrid data~\cite{chen-etal-2021-finqa}, which is a rich area for applications such as financial analysis. As demonstrated in Figure \ref{fig:intro}, a novel pipeline~\cite{chen-etal-2021-finqa} first uses a retriever to extract a subset of supporting evidence that contains the required variables from a long Table-and-Text hybrid passage. Then, a sequence-to-sequence program solver is trained on the retrieved variables to predict the solution program, as shown in the example of `$divide(1760, add(279,320))$'. This solution program can be represented as an abstract syntax tree, as illustrated in the right section of Figure \ref{fig:intro}.

Such retrieve-then-solve framework, while simple, has limitations that make it difficult for the model to learn the task effectively. One issue is that the retrieved evidence could be noisy and contain irrelevant variables, which hinders the performance of the program solver.  As illustrated in Figure \ref{fig:intro}, the red-colored variables `$170.1$' and `$7$' are irrelevant to the question. Numerical reasoning demands high precision, and systems are sensitive to noise. Studies on adversarial attacks for numerical question answering~\cite{kumar-etal-2021-adversarial-examples, patel-etal-2021-nlp, https://doi.org/10.48550/arxiv.2211.07455} have shown that irrelevant information can mislead the model and harm performance. The model may struggle to select the correct variables for reasoning, leading to incorrect predictions. Another limitation is that the program generation task requires the model to perform multiple steps of sub-program construction to generate the final solution program tree, which is a challenging task for deep learning systems~\cite{yang-etal-2018-hotpotqa}, especially considering the task involves table-specific operators that involve multiple variables such as \textit{`table\_average'}. Additionally, the system only receives coarse-grained supervision of the whole program during training, meaning it only knows the final program, not which evidence each sub-program is derived from. This makes it difficult for the model to learn about the underlying reasoning process and to generalize to new examples.

In this paper, we address these challenges and propose three auxiliary pretraining tasks. Specifically, to enhance the model's ability to distinguish between required and irrelevant variables in retrieved evidence, we propose the \textit{Variable Integrity Ranking} pretraining task. As illustrated in Figure \ref{fig:intro}, given two evidence and question pairs that contain different sets of variables such as orange \textit{`set 1'} that contains all required variables and green \textit{`set 2'} that only contains partial required variables, the model is trained to rank which set of variables contains more required variables. The model can learn to eliminate irrelevant variables and focus on useful supporting evidence for reasoning.

In order to provide fine-grained supervision for the construction of sub-programs, we propose two novel pretraining tasks, namely \textit{Variable Operator Prediction} and \textit{Variable Keyphrases Masking}, which decompose the coarse-grained annotation into sub-program level. We observe that the variables in the target program can provide direct guidance for sub-program construction by breaking down the program into the smallest sub-program unit consisting of two variables and an operator. As demonstrated in Figure \ref{fig:intro}, given an evidence and question pair, the model is trained to predict the operator (e.g., \textit{`add'}) between two variables (e.g., `$279$' and `$320$'), where the decomposition of the final program provides the supervision. This single-operator reasoning task helps the model learn the underlying reasoning of sub-program construction, thus allowing the model to perform more accurate numerical reasoning. 
We also observe that keyphrases such as technical terms and dates often serve as critical information for the construction of sub-program. As illustrated in Figure \ref{fig:intro}, given an evidence and question pair, the model is trained to recover the masked keywords describing the variables (e.g., \textit{`Units'}) by considering other pieces of evidence and inferring which variable information is required to answer the question. This allows the model to understand how keyphrases information determines the sub-program construction.

In the remaining of the paper, we introduce task settings and baseline model in Section \ref{sec:prel}, methodology of our three pretraining tasks in Section \ref{sec:method}, experimental settings of the dataset, evaluation metric and implementation details in Section \ref{sec:exp}, results of applying the three pretraining tasks to existing transformer-based pretrained language models (PLMs), ablation study and case study in Section \ref{sec:res}, and related works in Section \ref{sec:rel}.

\section{Preliminary Background}
\label{sec:prel}

As per the methodology outlined by \citet{chen-etal-2021-finqa}, for each example in the training data, a hybrid table-and-text hybrid passage, a question $Q$, a solution program $P$, and an annotation of the gold evidence set are provided. The cells in the table are transformed into sentences by concatenating the row and column headers using a template. For example, the purple cell labeled `Units' in Figure \ref{fig:intro} can be transformed into the sentence \textit{`The Charlotte at Midtown of Units is 279'}. The cells within the same row are concatenated to form a single piece of evidence.

We leverage FinQANet~\cite{chen-etal-2021-finqa} as the numerical reasoning model. It is composed of two main components: an evidence retriever that first retrieves the relevant evidence from the input financial report, followed by a program solver that generates the solution program.

\paragraph{Evidence Retriever} The original hybrid passage forms an evidence candidate sentence set $S = {s_1,...s_n}$, and the gold evidence forms a set of sentences $S^g = {s^g_1,...s^g_k}$. The objective of the evidence retriever is to assign a score to each sentence in $S$, which reflects the likelihood of it appearing in $S^g$. Each candidate supporting evidence sentence $s_i$ is concatenated with the question and then passed through a PLM classifier. The top-n retrieved facts, as determined by the scores of the classifier, are reordered as they appear in the input report and utilized as input for the program solver.

\paragraph{Program Solver} The program solver is an extension of sequence-to-sequence models, which takes the retrieved supporting evidence as the encoder input. The decoding target is a flattened representation of the solution program. For example, the program `$divide(1760, add(279,320))$' is flattened to $add(279,320), divide(1760, \#0)$, where $\#0$ denotes the first decoded sub-program $add(279,320)$. To enhance the model's ability of capturing structural information provided by the decoding history of sub-programs, the representations of these sub-programs are also provided as input to the decoder, in addition to the encoder outputs.

\section{Methodology}
\label{sec:method}

\begin{figure*}[t]
  \centering
  \includegraphics[width=0.8\linewidth]{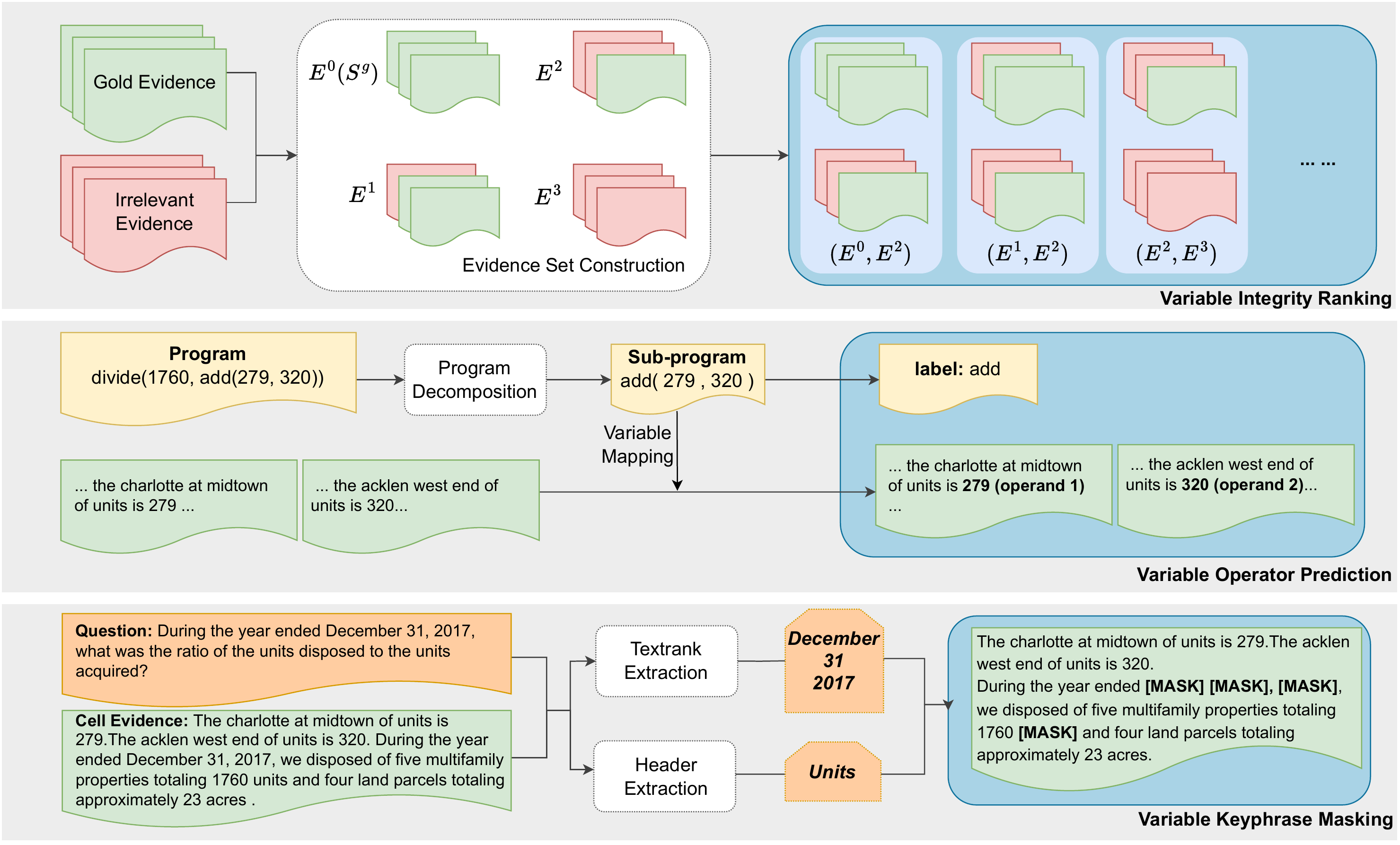}
  \caption{Pipeline of the construction of pretraining data. }
  \label{fig:frame}
  \end{figure*}

Our proposed approach applies the three pretraining tasks to a transformer-based PLM $Enc$, which serves as a universal encoder for the three tasks. The construction pipeline of the pretraining task data is depicted in Figure \ref{fig:frame}. In Section \ref{sub:nli}, we introduce \textit{Variable Integrity Ranking} (\textbf{VIR}), which utilizes whole program level supervision to guide the model to contextually learn which variables are required. In Section \ref{sub:re} and Section \ref{sub:mlm}, we present two sub-program level pretraining tasks: \textit{Variable Operator Prediction} (\textbf{VOP}), which leverages fine-grained supervision to enable the model to learn single operator reasoning, and \textit{Variable Keyphrase Masking} (\textbf{VKM}), which provides immediate guidance for the model to benefit from the reasoning of using keyphrases to construct sub-programs. We perform the pretraining using a multi-task training strategy, which is described in Section \ref{sub:multask}. 
We then use the pretrained model to initialize the FinQA retriever and solver encoder for task-specific fine-tuning.

\subsection{VIR: Variable Integrity Ranking}
\label{sub:nli}

The objective of this pretraining task is to train the model to rank the number of pieces of gold evidence included in a question and evidence sentence example, i.e., variable integrity levels. In order to construct examples with varying integrity levels, as depicted in Figure \ref{fig:frame}, sentences are selected from both gold (green) evidence and irrelevant (red) evidence within the passage to form pseudo evidence sets, denoted as $E^i$. The integrity level, represented as $i$, indicates the ranking by the number of gold pieces of evidence included. These generated sets are subsequently utilized for training the model in the learning-to-rank framework.

Specifically, given an example from the training data, the gold evidence set $S^g$ includes all the required variables and is thus defined as the level $0$ evidence set $E^0$. To construct subsequent evidence sets, one gold evidence $s^g_j$ is selected from $E^i$, and replaced with irrelevant evidence randomly drawn from the original passage $S$ that is not present in the gold evidence set $S^g$. This replacement process is repeated $k$ times, resulting in the collection of $k+1$ pseudo evidence sets ${E^0,...E^k}$. These sets are subsequently utilized for training the model in the learning-to-rank framework by creating evidence set pairs, denoted as ${(E^u, E^v)|_{v=u+1}^{k}|_{u=0}^{k-1}}$, yielding a total of $(k+1)*k/2$ pairs for pretraining.

In order to rank the integrity of variables, we employ a novel pairwise learning-to-rank algorithm, RankNet~\cite{burges2005learning}. The evidence set pair $(E^u, E^v)$ is separately concatenated with the question text $Q$ and fed into the pre-trained language model (PLM) $Enc$. The representations $(h^u, h^v)$ of the example are obtained by utilizing the \textit{[CLS]} token representations.

\begin{equation}
    h^u = Enc_{[CLS]}(Q;E^u)
\end{equation}

Then, the representations are mapped by feed-forward network (FFN) and activation function $tanh$ to a single dimension score $(s^u, s^v)$, which stands for the score of their variable integrity level $(u, v)$. The final loss is calculated by Binary Cross-Entropy over the subtraction of the two scores:
\begin{equation*}
\begin{aligned}
        s^u &= tanh(FFN(h^u))\\
        \mathcal{L}_{rank} &= BCE(s^u-s^v)
\end{aligned}
\end{equation*}

\subsection{VOP: Variable Operator Prediction}
\label{sub:re}

The proposed model is trained to predict the operator between two variables within a given question and evidence sequence. As illustrated in Figure \ref{fig:frame}, the program $divide(1760, add(279,320))$ can be decomposed into sub-programs $add(279,320)$ and $divide(1760, \#0)$. The latter sub-program contains a non-variable operand, for which the representation cannot be extracted from text. Thus to construct the training examples, the original solution program is first decomposed, and the sub-programs in which both operands are variables are extracted. The operand variables are then mapped to the row-based annotated gold evidence to identify the corresponding target variables.

The operator prediction task treats the operator $y_{op}$ as a relation between the operand variables and employs novel relation classification methods~\cite{bekoulis2018joint, zhong-chen-2021-frustratingly} to construct the operator prediction module. However, unlike relation classification tasks, some program functions may have more than two operands (e.g., \textit{table\_sum} can have three or more operands). So instead of concatenating the token representations, the final representation $h_{op}$ of the operator is calculated by averaging the PLM $Enc$ representations ${h_0, ... h_m}$ of the $m$ operands' first token. A feed-forward network (FFN) and softmax function map the representation to the operator prediction. The final loss is defined as:

\begin{equation*}
\begin{aligned}
    h_{op} &= avg(h_0,...h_m)\\
    \mathcal{L}_{op} &= NLL(softmax(FFN(h_{op}), y_{op})\\
\end{aligned}
\end{equation*}

Where $y_{op}$ belongs to \{\textit{`add', `subtract', `multiply', `divide',
          `exp', `greater', `table\_sum', `table\_average',
          `table\_max', `table\_min'}\}.

\subsection{VKM: Variable Keyphrase Masking}
\label{sub:mlm}

This task trains the model to recover masked keyphrases that describes the variables from a masked question and evidence sequence example. We leverage two keyphrase extraction methods. To effectively extract keyphrases that describe the required variables, we use cell-based evidence that contain less noise of irrelevant variables. 

First, we use TextRank~\cite{mihalcea-tarau-2004-textrank} over the concatenation of the question text and the gold evidence to extract keyphrases. 
TextRank~\cite{mihalcea-tarau-2004-textrank} constructs a graph over tokens considering co-occurrence. Then PageRank~\cite{page1999pagerank} algorithm is applied over the token graph to rank token importance. We only use the keyphrases that appear twice or more for masking, to ensure they are valid descriptions for the required variables.

Second, we consider a feature of table data, that headers naturally form keyphrases that describe the variables. However, not all headers could be reconstructed given the context, such as \textit{`The acklen west end'} in Figure \ref{fig:frame}, the name of this community is only given in the table row. We use the headers that appear twice or more for keyphrase masking, that these headers describe more than one variable, implying the relation of two or more variables.  

We follow the Masked Language Model pretraining of BERT~\cite{devlin-etal-2019-bert} to train the model to recover the keyphrases. Given the concatenation of question and cell-based evidence, each keyphrase is randomly masked for once occurrence. The masked input text is given to the transformer-based language model, and the model is trained to predict the token at the masked positions with Cross Entropy Loss $\mathcal{L}_{mask}$.

\subsection{Multi-task Pretraining}
\label{sub:multask}

To perform multi-task training, we leverage a streamlined version of MT-DNN~\cite{liu-etal-2019-multi,liu-etal-2020-microsoft}. We construct fixed mini-batches $\{b_t\}$, that in each mini-batch all examples are of the same pretraining task $t$. In each training step, a mini-batch $b_t$ is selected and the model is updated according to the task-specific loss $\mathcal{L}_t$ for the task $t$. This optimization procedure could be seen as an approximation of the sum of multi-task objectives $\mathcal{L}_{rank}$, $\mathcal{L}_{op}$ and $\mathcal{L}_{mlm}$.

\section{Experiments}
\label{sec:exp}

\subsection{Dataset}

In this study, we perform experiments utilizing the FinQA dataset~\cite{chen-etal-2021-finqa}. The FinQA dataset comprises 8,281 financial questions that were extracted from financial reports of companies of S\&P 500. The dataset is split into 6,251 examples for training, 883 examples for development validation, and 1,147 examples for testing. The maximum number of operators in a program is 6, with an average of 1.54 operators per program. The maximum number of gold evidence is 9, with an average of 1.71 pieces of evidence per program.

To the best of our knowledge, the TAT-QA dataset~\cite{zhu-etal-2021-tat} and the MultiHierr dataset~\cite{zhao-etal-2022-multihiertt} are the only other datasets that investigate hybrid table-and-text numerical reasoning. While their settings and textual domains are similar to that of FinQA, they contain span prediction questions in addition to program prediction questions. We consider FinQA since our proposal focuses on program solution prediction.

\subsection{Evaluation Metric}

 The overall performance of the dataset is evaluated using two metrics: execution accuracy (exe acc) and program accuracy (prog acc). Execution accuracy assesses the correctness of the final numerical value output of the program solution, while program accuracy evaluates the mathematical equivalence of the program solution to the gold program using a set of rules. Execution accuracy serves as an upper bound for system performance, while program accuracy serves as a lower bound.

 To examine the performance of the retriever, we report the top-3 recall (\textbf{R@3}) and top-5 recall (\textbf{R@5}) of the gold evidence. The recall metric denotes the proportion of successfully retrieved gold evidence out of the total gold evidence.

\subsection{Implementation Details}

Our method is implemented using Pytorch~\cite{NEURIPS2019_9015} and Transformers~\cite{wolf-etal-2020-transformers}. BERT-based models are trained on an NVIDIA 3090 GPU with 24G memory for approximately 12 hours, while Roberta-large models are trained on an NVIDIA A100 GPU with 40G memory for approximately 24 hours. For pretraining, we use a batch size of 4 and trained for 5 epochs. The initial learning rate is set to 5e-6, and we employ a warm-up proportion of 0.1.
For fine-tuning, we followed the hyperparameters of FinQANet~\cite{chen-etal-2021-finqa}, except for the learning rate. We observed that training failed to converge using the original settings on transformer-large PLMs, and thus add a cosine learning rate scheduler with 50 steps of warm-up, gradient clipping with a maximum norm of 1.0, and weight decay of 1e-5 to avoid gradient explosion. To prevent overfitting, we utilized loss-based early stopping with a patience of 30 epochs, and then selected the best model among the patience period for test using development set accuracy. We report results based on the average of three fixed random seeds. The top-3 retrieved evidence scored by the retriever is used as input for the program solver.

For the retriever, we do not apply sub-program level tasks, which are designed for sub-program construction. We observe that the retriever needs to handle a large volume of irrelevant information. To imitate the massive negative samples, we add an additional irrelevant evidence to the ranking sets during pretraining, namely noisy Variable Integrity Ranking for the retriever. For the program solver, we apply all three pretraining tasks.

\begin{table*}[t]
\centering
\begin{tabular}{lc|cccc}
\hline
\multirow{2}{*}{\textbf{Model}} & \multirow{2}{*}{\textbf{Solver PLM}} &\multicolumn{2}{c}{\textbf{Dev(\%)}}  & \multicolumn{2}{c}{\textbf{Test(\%)}} \\\cline{3-6}
 & & \textbf{exe acc} & \textbf{prog acc} & \textbf{exe acc} & \textbf{prog acc}\\
\hline
Retriever+NeRd&BERT-Base&23.83&22.56&48.57&46.76\\
Longformer&-&47.53&45.37&21.90&20.48\\
ELASTIC&Roberta-Large&-&-&62.16&57.54\\
\hline
FinQANet &BERT-Base&49.91&47.15&50.00&48.00\\
FinQANet$\P$ &BERT-Base&52.10&49.83&51.43&49.26\\
Ours &BERT-Base&55.61 & 52.77 & 55.80 & 53.44 \\
FinQANet &Roberta-Large & 61.22&58.05 &61.24 &58.86\\
FinQANet$\P$ &Roberta-Large &63.19&61.11&61.95&59.81\\
Ours &Roberta-Large&\textbf{67.04} & \textbf{63.76} & \textbf{65.51} & \textbf{63.55}\\
\hline
Human Expert&-&-&-&91.16&87.49\\
General Crowd&-&-&-&50.68&48.17\\
\hline

\end{tabular}
\caption{Main results of our method and baselines. \textbf{Dev} denotes performance on the development set. \textbf{Test} denotes performance on the test set. $\P$ denotes our implementation. \textbf{Solver PLM} denotes the PLM used for program solver.}
\label{tab:main}
\end{table*}
\begin{table}[t]
\centering
\begin{tabular}{l|cccc}
\hline
\multirow{2}{*}{\textbf{Model}} &\multicolumn{2}{c}{\textbf{Dev}}  & \multicolumn{2}{c}{\textbf{Test}}\\\cline{2-5}

 & \textbf{R@3} & \textbf{R@5} & \textbf{R@3} & \textbf{R@5}  \\
\hline
BERT-Base &90.47&93.91&88.98&93.56\\
VIR &90.97&94.22&89.28&\textbf{94.08}\\

$\mathcal{N}$VIR &\textbf{91.66}&\textbf{95.12}&\textbf{90.05}&93.66\\

\hline
\end{tabular}
\caption{Results of retriever performance on devlopment set and test set. \textbf{R@n} denotes the top-$n$ retriever recall of gold evidence. \textbf{VIR} denotes using Variable Integrity Ranking as pretraining task. \textbf{$\mathcal{N}$VIR} denotes using noisy Variable Integrity Ranking as pretraining task.}
\label{tab:retriever}
\end{table}

\section{Results and Analysis}
\label{sec:res}

\subsection{Main Results}

The experiment results are shown in Table \ref{tab:main}. We compare our method with various competitive methods. \footnote{There was a data-leakage in the early version of FinQA datasets, all results reported in this paper remove this bug and could exist
discrepancies with the results in original paper.}  \textbf{Retriever+NeRD}~\cite{ran-etal-2019-numnet} assigns plus, minus or zero to all variables and sums the signed variables. \textbf{Longformer}~\cite{Beltagy2020Longformer} uses a PLM designed for long text as the encoder and takes in all evidence text as the input. \textbf{ELASTIC}~\cite{zhangelastic} uses an adapted program solver that separates the generation of operators and operands; \textbf{General Crowd} and \textbf{Human Expert} refers to crowd workers and professional expert human performance reported in \citet{chen-etal-2021-finqa}, where the latter result serves as an upper bound of systems. \textbf{FinQANet} is our baseline model as described in section \ref{sec:prel}, we also provide results of our re-implementation. 

Following FinQANet, we use \textbf{BERT-base}~\cite{devlin-etal-2019-bert} for the retriever and give the comparison of our method and the baseline on two PLMs for the program solver: \textbf{BERT-Base} and \textbf{RoBERTa-Large}~\cite{https://doi.org/10.48550/arxiv.1907.11692} to demonstrate the effectiveness of our method on different scales of PLMs and different pretraining strategies. Our method surpasses all baselines and achieves the best performance on both PLMs. Specifically, on FinQA test set we achieve 4.37\% execution accuracy and 4.18\% program accuracy improvement with BERT-base program solver, and 3.56\% execution accuracy and 3.74\% program accuracy improvement with RoBERTa-Large program solver. 

These results demonstrate the effectiveness of our whole program and sub-program level pretraining approach. The Variable Integrity Ranking pretraining task can guide the model to recognize useful evidence, allowing the model to focus on the required variables during reasoning. The Variable Operator Prediction and Variable Keyphrase Masking pretraining tasks provide fine-grain supervision, which helps the model learn the underlying reasoning and perform better sub-program construction.

\subsection{Ablation Studies}
To further understand the performance of our method, we conducted an ablation study to break down the performance of each component.

\paragraph{Retriever Recall} 

 In Table \ref{tab:retriever}, we present a comparison of the retriever performance using different pretraining settings. Our proposed method outperforms the BERT-Base baseline on all evaluation metrics under both settings of Variable Integrity Ranking. Specifically, \textbf{$\mathcal{N}$VIR} achieves the best top-3 recall on both the development set and test set, gaining 1.19\% points and 1.07\% points of improvement, respectively. Additionally, \textbf{VIR} outperforms the baselines and achieves the best top-5 recall performance on the test set. These results demonstrate that our whole program level pretraining task is beneficial for the retriever, as it helps the model learn to distinguish irrelevant evidence. Furthermore, the noisy Variable Integrity Ranking setting further enhances the model's ability to extract evidence from massive negative irrelevant examples, although this advantage diminishes as the number of retrieved evidence increases.

\begin{table}[t]
\centering
\begin{tabular}{ll|cccc}
\hline
\multirow{2}{*}{\textbf{R}} & \multirow{2}{*}{\textbf{P-S}} & \multicolumn{2}{c}{\textbf{Dev(acc/\%)}}  & \multicolumn{2}{c}{\textbf{Test(acc/\%)}} \\\cline{3-6}
 &    & \textbf{exe} & \textbf{prog} & \textbf{exe} & \textbf{prog}\\
\hline
\multicolumn{6}{l}{\textit{BERT-base}}\\
\hline
\textendash & \textendash &52.10&49.83&51.43&49.26\\
$\mathcal{N}$VIR& \textendash & 53.00 & 50.85 & 52.22 & 49.87 \\
$\mathcal{N}$VIR&VIR & 54.81 & 52.22 & 53.96 & 51.22\\
$\mathcal{N}$VIR&VOP& 53.45 & 51.08 & 52.48 & 50.48\\
$\mathcal{N}$VIR&VKM& 52.10 & 49.60 & 52.57 & 50.13\\
$\mathcal{N}$VIR&ALL & \textbf{55.61} & \textbf{52.77} & \textbf{55.80} & \textbf{53.44}\\
 \hline
\multicolumn{6}{l}{\textit{Roberta-Large}}\\
 \hline

\textendash&\textendash &63.19&61.11&61.95&59.81\\
$\mathcal{N}$VIR&\textendash& 65.57 & 62.40 & 63.38 & 61.38 \\
$\mathcal{N}$VIR&VIR & 66.59 & 63.65 & 64.87 & 62.86\\
$\mathcal{N}$VIR&VOP & 66.02 & 62.51 & 63.99 & 61.63 \\
$\mathcal{N}$VIR&VKM & 65.46 & 62.51 & 64.60 & 62.25\\
$\mathcal{N}$VIR&ALL & \textbf{67.04} & \textbf{63.76} & \textbf{65.51} & \textbf{63.55}\\
\hline

\end{tabular}
\caption{Ablation study using different pretraining tasks. \textbf{R} denotes retriever and \textbf{P-S} denotes program solver. \textbf{\textendash} denotes the PLM baselines with no additional pretraining. \textbf{VOP}/\textbf{VKM} denotes using Variable Operator Prediction/Variable Keyphrase Masking as pretraining task. \textbf{All} denotes using all three pretraining tasks.}
\label{tab:ab}
\end{table}

\paragraph{Program Prediction Accuracy} 
To investigate how the pretraining tasks benefit the final program prediction, 
we conducted an ablation study evaluating the overall execution accuracy and program accuracy performance. As shown in Table \ref{tab:ab}, our method substantially improves program prediction results. Specifically, the retriever \textbf{$\mathcal{N}$VIR} on the test set gains 0.79\% execution accuracy and 0.61\% program accuracy with BERT-Base as the program solver, and 1.43\% execution accuracy and 1.57\% program accuracy with RoBERTa-Large as the program solver. The improvement in retriever recall, resulting from the whole program level pretraining, is directly reflected in the final prediction accuracy.
For the program solver, we examine the performance of using the three pretraining tasks separately, to verify their individual effectiveness. We observed that all three pretraining tasks improve the performance compared to using no auxiliary pretraining tasks. Among the three tasks, \textbf{VIR} achieves the most improvement, 1.74\% execution accuracy and 1.35\% program accuracy improvement on BERT-base, and 1.49\% execution accuracy and 1.48\% program accuracy improvement on RoBERTa-Large. These results show that distinguishing the required and irrelevant variables is crucial for the task. Additionally, the two sub-program level tasks also improved the test set accuracy of the program solver on both PLMs, demonstrating the effectiveness of fine-grain level supervision. We achieved the best results on both PLMs applying all three tasks on the program solver.

\begin{table}[t]
\centering
\begin{tabular}{lccc}
\hline
\textbf{Model} & VIR & VOP & VKM \\
\hline
VIR &93.78&-&-\\
VOP &-&92.70&-\\
VKM &-&-&52.43\\
All &94.91&93.57&54.41\\
\hline
\end{tabular}
\caption{Results of pretraining tasks accuracy.}
\label{tab:pre}
\end{table}
\subsection{Pretraining Task Performance}

 To investigate the model's ability to fit the objectives of the auxiliary tasks, we examined the RoBERTa-Large encoder performance of the three pretraining tasks on the FinQA development set. As shown in Table \ref{tab:pre}, the model is able to fit the objectives of the three tasks, achieving high accuracy of 94.91\% for Variable Integrity Ranking and 92.70\% for Variable Operation Prediction. We observed that multi-task pretraining can benefit the results of each individual task. These results demonstrate that jointly training the three tasks can further improve the model's ability to examine required variables at the whole program level and construct sub-programs at the sub-program level.

\subsection{Case Study}

\begin{figure*}[t]
  \centering
  \includegraphics[width=0.95\linewidth]{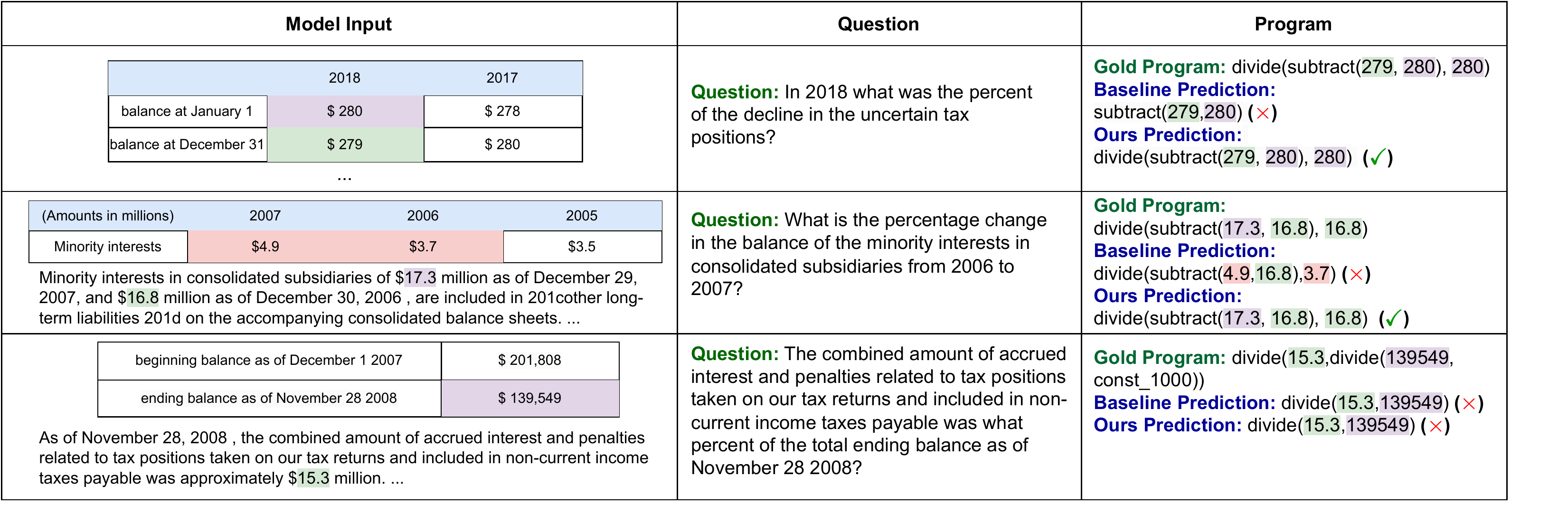}
  \caption{Case study on FinQA test set. We abbreviate unused model input evidence.}
  \label{fig:case}
  \end{figure*}

As shown in Figure \ref{fig:case}, we conducted a case study to further investigate how our method improves the whole program and sub-program reasoning of the model. We compared the results of the original solver and our proposal using RoBERTa-Large, given the same $\mathcal{N}$VIR retriever inputs. In case 1, while the baseline found the required variables, it failed to construct the \textit{`divide'} sub-program. However, our method, benefiting from fine-grain supervision, successfully predicted the correct program. In case 2, although the baseline model predicted the correct equation skeleton, it did not capture the information of \textit{`consolidated subsidiaries'} and was misled by the irrelevant \textit{`Minority Interests'} evidence in the model input when selecting variables. The Variable Integrity Ranking guides the model to focus on the required variables and generate the correct solution program. In case 3, although both systems captured the correct variable and operator information, they could not perform unit conversion of \textit{`million'} and \textit{`percentage'} and failed the prediction. The model needs further guidance to understand numerical concepts such as unit conversion, which we consider as future work.

\section{Related Works}
\label{sec:rel}

\subsection{Numerical Reasoning}

Recent studies of numerical reasoning have two major research lines. Machine Reading Comprehension (MRC) style tasks predicts a span or a option from multi-choice, such as DROP~\cite{dua-etal-2019-drop} which predicts a text or number span out of Wikipedia documents and questions and NumGLUE~\cite{mishra-etal-2022-numglue} that evaluates eight tasks that require simple arithmetic understanding. Solution prediction style tasks predicts a semantic parsing style program such as MathQA~\cite{amini-etal-2019-mathqa}, a mathematical equation such as Math23K~\cite{wang-etal-2017-deep} and MAWPs~\cite{koncel-kedziorski-etal-2016-mawps}, or a textual rational description of the reasoning process~\cite{cobbe2021training}. 
For tabular numerical reasoning, HiTab~\cite{cheng-etal-2022-hitab} dataset provides numerical reasoning question and program pairs on hierarchical tables.

For Table-and-Text Hybrid numerical reasoning, FinQA~\cite{chen-etal-2021-finqa}, TAT-QA~\cite{zhu-etal-2021-tat} and MultiHierr~\cite{zhao-etal-2022-multihiertt} consider numerical reasoning on financial reports. All questions in FinQA require program solution prediction, while a proportion of question in TAT-QA and MultiHierr require span prediction.

\subsection{Pretraining for Question Answering}

Various post-pretraining methods have been proposed for question answering. Span prediction tasks are utilized to benefit MRC~\cite{glass-etal-2020-span,ram2021few, deng-etal-2021-reasonbert, jiang-etal-2022-omnitab} which also use token masking recovery. They aim to use cloze-like data to learn MRC, while our approach focuses on fine-grain level sub-tree construction.
To enhance numerical reasoning skills of MRC systems, \citet{geva2020injecting} and \citet{https://doi.org/10.48550/arxiv.2201.11473} pretrains the model to generate code programs and equations. \citet{feng-etal-2021-pretraining-numerical} maps entity representations with numbers representations for Knowledge-Graph question answering. 

The closest studies to our method are MWP-BERT\cite{liang-etal-2022-mwp} and FORTAP~\cite{cheng-etal-2022-fortap} that consider solution prediction style numerical reasoning. Both studies leverage a series of pretraining tasks to understand value magnitude and predict components of the solution program, and the most similar tasks to our proposal is operator prediction. However, their studies limits to only textual or only tabular data, while our proposal considers the challenging hybrid table-and-text setting.

\section{Conclusion}

In this paper, we propose three solution program centric auxiliary pretraining tasks at both the whole program level and sub-program level. At the whole-program level, we propose the Variable Integrity Ranking pretraining task, which guides the model to distinguish required and irrelevant variables in the noisy input. To further enhance the model's ability to learn the underlying reasoning process, we propose two additional pretraining tasks: Variable Operator Prediction and Variable Keyphrase Masking. These tasks help the model perform accurate sub-program construction. Experimental results demonstrate the effectiveness of our method. Variable Integrity Ranking achieves the most improvement on both the retriever and program solver. The sub-program level tasks substantially improve results on PLMs of different scales. Our approach achieves 3.56\% execution accuracy and 3.74\% program accuracy improvement on the competitive RoBERTa-Large baseline.

\section*{Limitations}

Our pipeline has limitations in the following two aspects that we plan to address in the future:

\paragraph{Rely on Supervised Data} Despite the effectiveness, our method relies on expensive human expert annotated supervised data to perform the pretraining, while rich unlabeled resource of financial reports could be obtained. Various studies have investigated semi-supervised or self-supervised objectives for general domain tabular question answering~\cite{herzig-etal-2021-open} and textual MRC~\cite{pi-etal-2022-towards}, however designing a pretraining task for numerical reasoning on hybrid table-and-text data still remains a challenge, which we would investigate in future work.

\paragraph{Pretraining for Constant Number Concepts} As we show in case study, PLM encoders still struggle with handling basic numerical knowledge concepts such as unit conversion, while such information is given explicitly as constant numbers in the program. We neglect such constant number information during Variable Operator Prediction pretraining because it is difficult to obtain the contextual representation of the constant numbers. For future work, we would use these constant annotation to help the model understand number magnitude and unit conversion.

\bibliography{anthology,custom}
\bibliographystyle{acl_natbib}

\end{document}